\documentclass{sigir98}
\usepackage{url}
\usepackage{epsf}

\title{Japanese Probabilistic Information Retrieval\\ Using Location and Category Information}

\author{
\bf Masaki Murata, Qing Ma, Kiyotaka Uchimoto, \\
\bf Hiromi Ozaku, Masao Utiyama and Hitoshi Isahara\\[0.3cm]
Intelligent Processing Section, Kansai Advanced Research Center, \\
Communications Research Laboratory, Ministry of Posts and Telecommunication\\
588-2, Iwaoka, Nishi-ku, Kobe, 651-2492, Japan\\
Phone: +81-78-969-2181  Fax: +81-78-969-2189  E-mail: murata@crl.go.jp\\
http://www-karc.crl.go.jp/ips/murata\\[0.5cm]
{\bf Abstract}\\[0.5cm]
\begin{minipage}[h]{13cm}
\small
Robertson's 2-poisson information retrieve model 
does not use location and category information. 
We constructed a framework using 
location and category information in a 2-poisson model. 
We submitted two systems based on this framework  
to the IREX contest, Japanese language 
information retrieval contest held in Japan in 1999. 
For precision in the A-judgement measure they scored 
0.4926 and 0.4827, 
the highest values among the 15 teams and 22 systems 
that participated in the IREX contest. 
We describe our systems and 
the comparative experiments done 
when various parameters were changed. 
These experiments 
confirmed the effectiveness of 
using location and category information. \\[0.5cm]
{\bf Keyword:} 2-poisson model, Location information, Category information
\end{minipage}\vspace{0.5cm}
}

\def\None#1{}

\def\Small#1{}

\def\OK#1{}

\def\Out#1{}

\def\it#1{}

\begin{document}

\thispagestyle{plain}
\setlength{\footskip}{1cm}
\pagestyle{plain}

\maketitle

\None{
\begin{abstract}
Robertson's 2-poisson information retrieve model 
does not use location and category information. 
We constructed a framework using 
location and category information in a 2-poisson model. 
We submitted two systems based on this framework  
to the IREX contest, Japanese language 
information retrieval contest held in Japan in 1999. 
For precision in the A-judgement measure they scored 
0.4926 and 0.4827, 
the highest values among the 15 teams and 22 systems 
that participated in the IREX contest. 
We describe our systems and 
the comparative experiments done 
when various parameters were changed. 
These experiments 
confirmed the effectiveness of 
using location and category information. 
\end{abstract}
\section*{Keywords}
2-poisson model, Location information, Category information
}

\section{Introduction}
Information retrieval (IR) 
has become an increasingly important area of research 
due to the rapid growth of the Internet. 
In 1999 the Information Retrieval and Extraction Exercise contest 
(IREX) was held in Japan. 
We submitted two systems to this contest. 
Their precision in the A-judgement measure\footnote{
A-judgment means that 
a document whose topic is relevant to a query 
is judged a relevant document.} was 
0.4926 and 0.4827, 
the highest values among the 15 teams and 22 systems 
in the IREX contest. 
This paper describes our systems and 
the comparative experiments done when various parameters were changed. 

Our information retrieval method uses 
Robertson's 2-poisson model \cite{2poisson}, which is one kind of 
probabilistic approach. But, Robertson's method 
does not use location or category information, 
which should be used to facilitate information retrieval. 
Against this background, 
we constructed a framework by using 
location information, category information, and 
detailed information in a 2-poisson model\footnote{The reason 
that this paper is entitled 
``Probabilistic Information Retrieval 
Using Location Information and Category Information'' is that 
our methods use location and category information 
in addition to Robertson's probabilistic information retrieval method.}. 
We verified the effectiveness of using 
these three types of information by doing comparative experiments. 
When the 2-poisson model is used 
a term extraction method needs to be selected. 
In this paper, we describe four term extraction methods, 
and compared them in experiments. 

\section{Information retrieval}

\subsection{Task}

The information retrieval tasks in this paper are 
identical to those for the IREX contest. 
The database used for information retrieval 
(the same used in IREX) is from 
two-years (1994-1995) of a Japanese newspaper \cite{mainichi}. 
We retrieved from this database 
documents which satisfied the information 
condition for a Japanese language query. 

The following is an example of a query. 
(The data is from the IREX preliminary experiment.) 

\vspace{0.1cm}

\begin{quote}
\baselineskip=0.97\baselineskip
{\bf Example of a Japanese query}
\begin{verbatim}
<TOPIC>
<TOPIC-ID>1001</TOPIC-ID>
<DESCRIPTION>企業合併</DESCRIPTION>
<NARRATIVE>記事には企業合併成立の発表が述べられており、その合併に参加する企業の名前が認定できる事。また、合併企業の分野、目的など具体的内容のいずれかが認定できる事。企業合併は企業併合、企業統合、企業買収も含む。</NARRATIVE>
</TOPIC>
\end{verbatim}

{\bf English translation}
\begin{verbatim}
<TOPIC>
<TOPIC-ID>1001</TOPIC-ID>
<DESCRIPTION>enterprise amalgamation
</DESCRIPTION>
<NARRATIVE>
The condition for a relevent document is 
that in the document an announcement of 
enterprise amalgamation materialization is 
described and the name of the enterprise 
which participated the amalgamation can be 
recognized. Also, one of the field of 
amalgamation enterprise and its purpose 
should be able to recognized. Enterprise 
amalgamation contains enterprise annexation, 
enterprise integration, and enterprise 
purchasing. </NARRATIVE>
</TOPIC>
\end{verbatim}
\end{quote}

\vspace{0.1cm}

The number indicated by \verb+<TOPIC-ID>+ means the ID number of the query. 
\verb+<DESCRIPTION>+ contains 
a phrase that indicates the information needed. 
\verb+<NARRATIVE>+ contains 
the sentences that restrict the information requested. 
During the task,  
the system receives a query such as the above one and 
outputs 300 documents in order of confidence. 

\vspace{0.1cm}

\begin{quote}
\baselineskip=0.97\baselineskip
{\bf Example of a Japanese document}
\begin{verbatim}
<DOCNO>950217091</DOCNO>
<SECTION>経済</SECTION>
<HEADLINE>キグナス石油精製を東燃が１００％子会社化</HEADLINE>
<TEXT>
　東燃は十六日、系列のキグナス石油精製（資本金十億円、本社・川崎市、森利英社長）を一〇〇％子会社化すると発表した。同社は東燃が七割、ニチモウが三割出資しており、東燃はニチモウが所有する全株式六十万株を百二十五億円で買収する。
</TEXT>
</DOC>
\end{verbatim}

{\bf English translation}
\begin{verbatim}
<DOCNO>950217091</DOCNO>
<SECTION>Economic page</SECTION>
<HEADLINE>Tounen company completely makes 
Kigunasu company a subsidiary company
</HEADLINE>
<TEXT>
Tounen company anounced that it makes 
Kigunasu comapry, one of its group companies, 
(The capital is one billion yen. The head 
office is in Kawasaki city. The president 
is Mr. Toshihide Mori.) a completely 
subsidiary company. Kigunasu company is 
invested 70% by Tounen company and 30% by 
Nichimou company, and Tounen company 
purchases 600,000 stocks which Nichimou 
company possesses at 12,500,000,000 yens. 
</TEXT>
</DOC>
\end{verbatim}
\end{quote}

\vspace{0.1cm}

In this document, 
the newspaper information category 
(the economic or political pages) is 
indicated by \verb+<SECTION>+, 
the title of the document is indicated by \verb+<HEADLINE>+, 
and the body of the document is indicated by \verb+<TEXT>+. 
The tool ``trec\_eval'' of TREC is used 
to evaluate the retrieval results \cite{trec_eval}. 
In the contest, 
``R-Precision'' were used. 
It indicates the precision when 
retrieving R documents, 
where R is the number of relevant documents. 

\subsection{Outline of our method}
\label{sec:robertson}

Our information retrieval method uses 
Robertson's 2-poisson model \cite{2poisson} which is one kind of 
probabilistic approach. 
Robertson's method calculates 
each document's score using the following equation\footnote{This equation 
is BM11, which corresponds to BM25 in the case of $b = 1$\cite{robertson_trec3}. 
Although we made experiments testing some cases of $b$ in BM25, 
the case of $b = 1$ was roughly better than any other cases 
in this work. So we used BM11.}, 
then outputs the documents with high scores as retrieval results. 
(The following $Score(d,q)$ is the score of a document $d$ 
against a query $q$.) 

{\footnotesize
\begin{eqnarray}
  \label{eqn:robertson}
&  Score(d,q) = \displaystyle \sum_{\begin{minipage}[h]{0.8cm}
    term $t$ 
    
    in $q$
    \end{minipage}} & \hspace*{-0.3cm} \left(  \displaystyle \frac{tf(d,t)}{\displaystyle tf(d,t) + k_{t} \frac{length(d)}{\Delta}} \ × \ log\frac{N}{df(t)} \right. \nonumber \\
& &  \left. × \displaystyle \frac{tf_{q}(q,t)}{tf_{q}(q,t) + kq} \right)
\end{eqnarray}}
where terms occur in a query. 
$tf(d,t)$ is the frequency of a term $t$ in a document $d$, 
$tf_{q}(q,t)$ is the frequency of $t$ in a query $q$, 
$df(t)$ is the number of the documents in which $t$ occurs, 
$N$ is the total number of documents, 
$length(d)$ is the length of a document $d$, and 
$\Delta$  is the average length of the documents. 
$k_{t}$ and $k_{q}$ are constants which are set by experiments. 

In this equation, 
we call {\footnotesize $ \frac{tf(d,t)}{\displaystyle tf(d,t) + k_{t} \frac{length(d)}{\Delta}}$}
the TF term, (abbr. $TF(d,t)$), 
{\footnotesize $log\frac{N}{df(t)}$} the IDF term, (abbr. $IDF(t)$), 
and {\footnotesize $\frac{tf_{q}(q,t)}{tf_{q}(q,t) + kq}$}
the TF$_{q}$ term, (abbr. $TF_{q}(q,t)$). 


Our method adds several extended terms 
to this equation, and is expressed by the following equation. 

{\footnotesize
\begin{eqnarray}
&& Score(d.q)  = \displaystyle K_{\tiny category}(d) \left\{ \displaystyle \sum_{\begin{minipage}[h]{0.8cm}
      term $t$

      in $q$
\end{minipage}} ( TF(d,t) \ × \ IDF(t)  \right. \nonumber \\
&&\hspace*{0.7cm} × TF_{q}(q,t) × \ K_{\tiny detail}(d,t) \ × \ K_{\tiny location}(d,t) ) \nonumber \\
&&\hspace*{0.7cm} \left. + \displaystyle \frac{length(d)}{length(d) + \Delta} \right\}
  \label{eqn:score}
\end{eqnarray}}

The TF, IDF and TF$_{q}$ terms in this equation are 
identical to those in Eq. (\ref{eqn:robertson}). 
The term {\footnotesize $\frac{length}{length + \Delta}$} 
has a higher value when a document is longer. 
This term is made because 
if the other information is exactly equal, 
the longer document is more likely to include 
the content requested by the query. 
$K_{\tiny category}$, $K_{\tiny detail}$ and $K_{\tiny location}$ 
are extended numerical terms made to improve precision. 
$K_{\tiny category}$ uses the category information of the document 
found in newspapers, such as 
the economic and 
political pages. 
$K_{\tiny location}$ uses the location of the term in the document. 
If a term is in the title or 
at the beginning of the body of the document, 
it is given a higher weighting. 
$K_{\tiny detail}$ uses the information 
such as whether 
the term is a proper noun and 
or a stop word such as 文書 ``document'' and もの ``thing''. 
In the next section,
we explain these extended numerical terms in detail. 

\subsection{Extended numerical terms}

We use the three extended numerical terms of
$K_{\tiny location}$, $K_{\tiny category}$, and $K_{\tiny detail}$ 
as in Eq. (\ref{eqn:score}). 
This section explains them in detail. 

\begin{enumerate}
\item 
\underline{Location information ($K_{\tiny location}$)} 

In general, 
the title or the first sentence of the body of a document 
in a newspaper very often indicates its subject. 
Therefore, the precision of information retrieval 
can be improved by weighting the terms in these two locations. 
The term $K_{\tiny location}$ performs this task, and 
changes the weight of a term 
based its location at the beginning of the document. 
If a term is in the title or 
at the beginning of the body, 
it is given a high weighting. 
Otherwise, it is given low weighting. 
This term is expressed as follows: 

{\footnotesize
\begin{eqnarray}
\hspace*{-0.5cm}
K_{\tiny location}(d,t) = \left\{ 
  \begin{array}[h]{l}
k_{\tiny location,1}  \\
\mbox{(when a term $t$ occurs in the title of}\\
\mbox{a document $d$),}\\[0.2cm]
1 + k_{\tiny location,2} \displaystyle \frac{(length(d) - 2*P(d,t))}{length(d)}\\
\mbox{(otherwise)}
  \end{array}\right.\hspace*{-0.5cm}
\label{eqn:ichi}
\end{eqnarray}}

$P(d,t)$ is the location where a term $t$ occurs 
in the document $d$. 
When a term occurs more than once in a document, 
its first occurrence is used. 
$k_{\tiny location,1}$ and $k_{\tiny location,2}$ are 
constants which are set by experiments. 

\item 
\underline{Category information ($K_{\tiny category}$)} 

$K_{\tiny category}$ uses category information 
such as the economic and political pages. 
This functions as a technique called relevance feedback \cite{r-feedback}. 
First, we specify the categories which occur 
in the top 100 documents of the first retrieval 
when $K_{\tiny category} = 1$. 
Then, we increase the scores of documents having the same categories. 
For example, 
if economic pages often occur in the top 100 documents 
of the first retrieval, 
we increase the score of a document whose page is 
a economic page 
and decrease the score of the document whose page is different. 
$K_{\tiny category}$ is expressed as follows; 

\begin{equation}
\footnotesize
  \label{eqn:men}
\displaystyle K_{\tiny category}(d) = 1 + k_{\tiny category} \frac{(Ratio A(d) - Ratio B(d))}{(Ratio A(d) + Ratio B(d))}
\end{equation}
where $Ratio A$ is the ratio of a category in the top 100 documents 
of the first retrieval. 
$Ratio B$ is the ratio of a category in all the documents. 
The value of $K_{\tiny category}(d)$ is large, 
when $Ratio A$ is large 
(page of a document $d$ occurs frequently 
in the top 100 documents of the first retrieval.) 
and $Ratio B$ is small 
(page of a document $d$ does not occur often 
in all the documents.). 
$k_{\tiny category}$ is a constant which 
is set by experiments. 

\item 
\underline{Other information ($K_{\tiny detail}$)} 

$K_{\tiny detail}$ is a more detailed numerical term that 
uses different information, such as whether 
the term is a proper noun and 
whether the term is a stop word such as 
文書 ``document'' and もの ``thing''.  
If a term is a proper noun, 
it is weighted high. 
If a term is a stop word, such as 文書 ``document'' and もの ``thing,'' 
it is weighted low. 
$K_{\tiny detail}$ is expressed as follows 
for simplicity, 
the variables for a document and a term, 
$d$ and $t$, are omitted: 

{\footnotesize
\begin{eqnarray}
  \label{eqn:detail}
K_{\tiny detail} & = & K_{\tiny descr} × K_{\tiny proper} × K_{\tiny nado} × K_{\tiny num}  \nonumber\\
& & × K_{\tiny hira} × K_{\tiny neg} × K_{\tiny stopword} 
\end{eqnarray}}

Each term in this equation is explained below. 

\begin{itemize}
\item 
  $K_{\tiny descr}$

  When a term is obtained from 
  the title of a query, i.e. DESCRIPTION, 
  $K_{\tiny descr}$ =  $k_{\tiny descr} (> 1)$. 
  Otherwise, $K_{\tiny descr}$ = 1. 
  This is because 
  a term obtained from the title of a query 
  is important. 

\item 
  $K_{\tiny proper}$

  When a term is a proper noun, 
  $K_{\tiny proper}$ = $k_{\tiny proper} (> 1)$. 
  Otherwise $K_{\tiny proper}$ = 1. 
  This is because 
  a term that is a proper noun is important. 

\item 
  $K_{\tiny nado}$

  When a term is followed by the Japanese word {\em nado} (such as) 
  in a query sentence, 
  $K_{\tiny nado}$ = $k_{\tiny nado} (> 1)$. 
  Otherwise $K_{\tiny nado}$ = 1. 
  A term which is followed by the Japanese word {\em nado} 
  is specific in meaning 
  and is just as important as a proper noun. 

\item 
  $K_{\tiny num}$

  When a term is numeric, 
  $K_{\tiny num}$ = $k_{\tiny num} (< 1)$. 
  Otherwise, $K_{\tiny num}$ = 1. 
  A term which consists of only numerals 
  does not contain much relevant information 
  making it unimportant to a query. 

\item 
  $K_{\tiny hira}$

  When a term consists of {\em hiragana} characters only, 
  $K_{\tiny hira}$ = $k_{\tiny hira} (< 1)$. 
  Otherwise, $K_{\tiny hira}$ = 1. 
  A term which consists of only {\em hiragana} characters 
  does not contain much relevant information 
  making it unimportant to a query. 

\item 
  $K_{\tiny neg}$

  When a term is obtained from a region tagged with a NEG tag in a query, 
  $K_{\tiny neg} = k_{\tiny neg}$. 
  Otherwise $K_{\tiny neg} = 1$. 
  
  In a query of the IREX contest, 
  an expression, ``... {\em wa nozoku}'' (... is excepted), 
  as in the following query, 
  was tagged with a NEG tag. 

\vspace{0.1cm}

  \begin{quote}
\baselineskip=0.97\baselineskip
\mbox{\bf Example Japanese query}
\begin{verbatim}
<TOPIC>
<TOPIC-ID>1003</TOPIC-ID>
<DESCRIPTION>国連軍の派遣</DESCRIPTION>
<NARRATIVE>平和維持活動など国連の活動における国連軍の派遣について述べられている記事。派遣の目的または対象地域が記事から明示的に分る事。<NEG>日本の自衛隊を国連に派遣するかどうかという問題のみに関する記事は除く。</NEG></NARRATIVE>
</TOPIC>
\end{verbatim}

\vspace{0.1cm}

{\bf English translation}

\vspace{0.1cm}

\begin{verbatim}
<TOPIC>
<TOPIC-ID>1003</TOPIC-ID>
<DESCRIPTION>Dispatch of the United 
Nations forces</DESCRIPTION>
<NARRATIVE>The condition for a 
relevent document is that in the 
document a dispatch of the United 
Nations forces in the activity of UN 
such as peace maintainence activity 
is described. The purpose of the 
dispatch or the target region should 
be described. <NEG>A document 
describing the discussion of whether 
the Self-Defense Forces of Japan is 
dispatched to UN or not is elimated.
</NEG></NARRATIVE>
</TOPIC>
\end{verbatim}
  \end{quote}

\vspace{0.1cm}

  If a term 
  from a region tagged with a NEG tag is used, 
  non-relevant documents are often retrieved 
  and therefore such a term is weighted low. 
  In this paper, 
  $k_{\tiny neg}$ is set to 0. 
  This indicates that 
  a term from a region tagged with a NEG tag is not used 
  in retrieval. 

\item 
  $K_{\tiny stopword}$

  When a term is a stopword such as 
  {\em jouken} (condition), {\em kiji} (document) and {\em baai} (case), 
  $K_{\tiny stopword}$ = $k_{\tiny stopword} (< 1)$. 
  Otherwise $K_{\tiny stopword}$ = 1. 
  A term that is a stopword 
  is unimportant. 

\end{itemize}

Each constant, such as $k_{\tiny descr}$, 
is set experimentally. 

\end{enumerate}

\subsection{How to extract terms}
\label{sec:extract_keyword}
Before being able to use Eq. (\ref{eqn:score}) in information retrieval, 
we must extract the terms from a query. 
This section describes how to do this. 
With regard to term extraction, 
we considered the several methods listed below. 

\begin{enumerate}
\item 
\underline{Method using only the shortest terms}

This is the simplest method. 
The method divides the query sentence into short terms 
by using the 
morphological analyzer ``juman'' \cite{JUMAN3.5e} 
and eliminates non-nominal words and stop words\footnote{Since, 
Japanese is an agglutinative languages like Chinese, 
there are no spaces between words and 
a morphological analyzer is necessary to divide a sentence into words.}. 
The remaining words are used in the retrieval process.

\None{
\begin{quote}
\baselineskip=0.97\baselineskip
\begin{verbatim}
<TOPIC>
<TOPIC-ID>1001</TOPIC-ID>
<DESCRIPTION>企業合併</DESCRIPTION>
<NARRATIVE>記事には企業合併成立の発表が述べられており、その合併に参加する企業の名前が認定できる事。また、合併企業の分野、目的など具体的内容のいずれかが認定できる事。企業合併は企業併合、企業統合、企業買収も含む。</NARRATIVE>
</TOPIC>
\end{verbatim}
\end{quote}}

\item 
\underline{Method using all term patterns}

In the first method 
the terms are too short. 
For example, 
\None{企業} ``enterprise'' and \None{合併} ``amalgamation'' 
are used instead of 
\None{企業合併} ``enterprise amalgamation.''\footnote{Although 
this paper deals only with Japanese, not English, 
for this explanation we use English examples 
for the English readers. 
This method handles compound nouns 
and can be used not only for Japanese 
but also for English. } 
We thought that we should use 
\None{企業合併} 
``enterprise amalgamation'' in addition to the two short terms. 
Therefore, we decided to use both short and long terms. 
We call this ``all-term patterns method.'' 
For example, when \None{企業合併成立} 
``enterprise amalgamation materialization'' was inputted, 
we use 
\None{企業} ``enterprise'', \None{合併} ``amalgamation'', 
\None{成立} ``materialization'', 
\None{企業合併} ``enterprise amalgamation'', 
\None{合併成立} ``amalgamation materialization'', 
and \None{企業合併成立} ``enterprise amalgamation materialization'' 
as terms for information retrieval. 
We thought that 
this method would be effective 
because it uses all term patterns. 
But, we also thought that 
it is inequitable that 
only the three terms 
of \None{企業} ``enterprise,'' 
\None{合併} ``amalgamation,'' 
\None{成立} ``materialization,'' 
are derived from \None{[...企業...合併...成立...]}
``... enterprise ... amalgamation ... materialization ...'',
while on the other hand 
six terms are derived from 
\None{企業合併成立} ``enterprise amalgamation materialization.'' 
We examined several normalization methods in 
preliminary experiments, 
and decided to divide the weight of each term 
by $\sqrt{\frac{n(n+1)}{2}}$, 
where $n$ is the number of successive words. 
For example, in the case of ``enterprise amalgamation materialization'', 
$n = 3$. 

\item \underline{Method using a lattice}

Although the method using all-term patterns is effective 
for use with all patterns of terms, 
it needs to be normalized by using 
the adhoc equation $\sqrt{\frac{n(n+1)}{2}}$. 
Thus, we considered 
the method where 
all the term patterns are stored into 
a lattice structure. 
We use the patterns in the path 
where the score in Eq. (\ref{eqn:score}) 
is the highest. 
(This method is almost same as 
Ozawa's \cite{ozawa_nlp99_eng}. 
The differences are 
the fundamental equation for information retrieval, 
and whether to use or not use a morphological analyzer.)

\begin{figure}[t]
  \begin{center}
  \begin{minipage}{7cm}
      \begin{center}
        \hspace*{-0.5cm}
        \epsfile{file=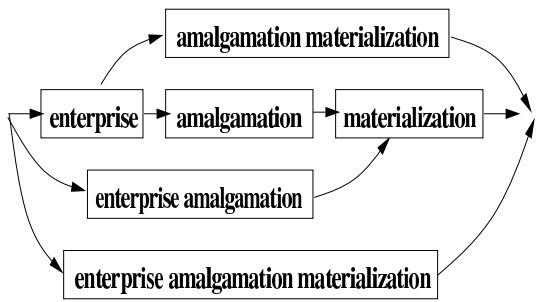,height=4cm,width=7.5cm} 
      \end{center}
      \vspace{-0.5cm}
    \caption{An example of a lattice structure}
    \label{fig:lattice}
    \end{minipage}
  \end{center}
\end{figure}

For example, 
in the case of ``enterprise amalgamation materialization'' 
a lattice, as shown in Fig. \ref{fig:lattice}, is 
obtained. 
As in this figure, 
four paths exist where 
each of their scores are calculated by Eq. (\ref{eqn:score})
and the terms in the highest path are used. 
This method does not require 
the adhoc normalization as in the method using all the term patterns. 

\item \underline{{Method using} down-weighting {\cite{Fujita99_IREX}}}

This is the method that Fujita proposed at the IREX contest, 
and we examined after the contest. 
It is similar to the all-term patterns method. 
It uses all the term patterns but 
the normalization is different from 
the all-term patterns method. 
It does not change the weight of the shortest terms; 
and decreases the weight of the longer terms. 
We decided to multiply the weight $k_{down}$$^{x-1}$ 
to a term, 
when it consisted of $x$ shortest terms, 
where $k_{down}$ was set by experiments. 
This method basically uses the shortest terms while 
also using the longer terms by down-weighting them. 

\end{enumerate}

\section{IREX contest results}
\label{sec:contest}

For our two submissions to the IREX contest\footnote{
IREX allowed two systems to be submitted.}, 
we selected the ``all-term patterns'' and 
``lattice structure'' methods to extract terms\footnote{
The reason we did not use 
``the shortest terms method'' 
is because it is too simple and did not 
seem effective. 
The ``down-weighting method'' is a method proposed at IREX. 
So we could not use it in IREX.}, 
and set the constants of the extended terms 
in order to maximize the precision 
in the preliminary-run data as follows. 

\begin{enumerate}
\item 
  System A

  It used the lattice method 
  for the term extraction. 
  The parameters were set as follows; 
  $k_{\tiny location,1} = 1.35$, 
  $k_{\tiny location,2} = 0.125$,  
  $k_{\tiny category} = 0$, 
  $k_{\tiny descr} = 1.5$,
  $k_{\tiny proper} = 2$,  
  $k_{\tiny nado} = 1$,  
  $k_{\tiny num} = 0.5$,  
  $k_{\tiny hira} = 0.5$,  
  $k_{\tiny neg} = 0$,  
  $k_{\tiny stopword,1} = 0$,  
  $k_{\tiny stopword,2} = 0.5$,  
  $k_{t} = 1$, and  
  $k_{q} = 0.1$. 
  Terms obtained from DESCRIPTION 
  are handled as terms different from 
  terms obtained from NARRATIVE. 

\item 
  System B

  It used  all-term patterns method 
  for term extraction. 
  The parameters were set as follows; 
  $k_{\tiny location,1} = 1.3$, 
  $k_{\tiny location,2} = 0.15$, 
  $k_{\tiny category} = 0.1$, 
  $k_{\tiny descr} = 1.75$, 
  $k_{\tiny proper} = 2$, 
  $k_{\tiny nado} = 1.7$, 
  $k_{\tiny num} = 0.5$, 
  $k_{\tiny hira} = 0.5$, 
  $k_{\tiny neg} = 0$, 
  $k_{\tiny stopword,1} = 0$, 
  $k_{\tiny stopword,2} = 0.5$, 
  $k_{t} = 1$, and 
  $k_{q} = 0$. 
  Terms obtained from DESCRIPTION 
  were handled as the terms different from 
  those obtained from NARRATIVE. 

\end{enumerate}
%


\begin{table}[t]
  \caption{R-Precision of all the systems}
  \label{tab:all_sys_result}
  \begin{center}
\begin{tabular}[c]{|l|c|c|}\hline
System ID & \multicolumn{1}{c|}{A-Judgment}  & \multicolumn{1}{c|}{B-Judgment} \\\hline
1103a  & 0.4505 & 0.4888 \\
1103b  & 0.4657 & 0.5201 \\
1106   & 0.2360 & 0.2120 \\
1110   & 0.3329 & 0.4276 \\
1112   & 0.2790 & 0.3343 \\
1120   & 0.2713 & 0.3339 \\
1122a  & 0.3808 & 0.4689 \\
1122b  & 0.4034 & 0.4747 \\
1126   & 0.0966 & 0.0891 \\
1128a  & 0.3384 & 0.3897 \\
1128b  & 0.3924 & 0.4175 \\
1132   & 0.0602 & 0.0791 \\
1133a  & 0.2383 & 0.2277 \\
1133b  & 0.2457 & 0.2248 \\
{\bf 1135a}  & 0.4926 & 0.5119 \\
{\bf 1135b}  & 0.4827 & 0.4878 \\
1142   & 0.4455 & 0.4929 \\
1144a  & 0.4658 & 0.5510 \\
1144b  & 0.4592 & 0.5442 \\
1145a  & 0.3352 & 0.3424 \\
1145b  & 0.2553 & 0.2935 \\
1146   & 0.2220 & 0.2742 \\\hline
\end{tabular}
\end{center}
\end{table}

In the contest, 
the results for the 22 systems were submitted 
by the 15 teams. 
Their R-Precisions are shown in Table \ref{tab:all_sys_result}. 
The first column of the table indicates 
the names of the systems. 
Our two systems, System A and System B 
correspond to 1135a and 1135b. 
A-Judgement and B-Judgement 
are the evaluation criteria determined by the IREX committee. 
A-Judgment means that 
a document whose topic is relevant to a query 
is judged as a relevant document. 
B-Judgment means that 
a document whose topic is partly relevant to a query 
is also judged as a relevant document. 
Although our systems were not the highest 
in B-Judgment, 
they were the highest among all the systems in A-Judgment. 
This result indicates that 
our method is relatively superior. 

\begin{table*}[t]
\footnotesize\renewcommand{\arraystretch}{1}
  \caption{Comparison of methods to extract keywords}
  \label{tab:result_keyward}
  \begin{center}

\mbox{(a) When all extended terms are used}
\begin{tabular}[c]{|l||l|l|l|l||l|l|l|l|}\hline
 & \multicolumn{4}{c||}{Formal run} & \multicolumn{4}{c|}{Preliminary run} \\\cline{2-9}
\multicolumn{1}{|c||}{Method to extract terms}  & \multicolumn{2}{c|}{R-Precision} & \multicolumn{2}{c||}{Average precision} & \multicolumn{2}{c|}{R-Precision} & \multicolumn{2}{c|}{Average precision} \\\cline{2-9}
\multicolumn{1}{|c||}{} & \multicolumn{1}{c|}{A-Judge}  & \multicolumn{1}{c|}{B-Judge} & \multicolumn{1}{c|}{A-Judge}  & \multicolumn{1}{c||}{B-Judge} & \multicolumn{1}{c|}{A-Judge}  & \multicolumn{1}{c|}{B-Judge} & \multicolumn{1}{c|}{A-Judge}  & \multicolumn{1}{c|}{B-Judge}  \\\hline 
Using the shortest terms   & 0.5012  & 0.5205$^{**}$  & 0.4935$^{**}$  & 0.4764$^{*}$  & 0.4412  & 0.5442  & 0.4546  & 0.5151\\
Using all term patterns$^{\#}$  & 0.4827 & 0.4878 & 0.4553 & 0.4453  & 0.4373  & 0.5573  & 0.4576  & 0.5317\\
Using the lattice structure  & 0.4926 & 0.5119 & 0.4808 & 0.4698  & 0.4599  & 0.5499  & 0.4638  & 0.5170\\
Using down-weight ($k_{down} = 0.01$) & 0.5006  & 0.5217  & 0.4935  & 0.4778  & 0.4412  & 0.5445  & 0.4546  & 0.5157\\
Using down-weight ($k_{down} = 0.1$)  & 0.4997  & 0.5233  & 0.4939  & 0.4809  & 0.4478  & 0.5504  & 0.4563  & 0.5185\\\hline
\end{tabular}

\mbox{(b) When no extended terms are used}
\begin{tabular}[c]{|l||l|l|l|l||l|l|l|l|}\hline
 & \multicolumn{4}{c||}{Formal run} & \multicolumn{4}{c|}{Preliminary run} \\\cline{2-9}
\multicolumn{1}{|c||}{Method to extract terms}  & \multicolumn{2}{c|}{R-Precision} & \multicolumn{2}{c||}{Average precision} & \multicolumn{2}{c|}{R-Precision} & \multicolumn{2}{c|}{Average precision} \\\cline{2-9}
\multicolumn{1}{|c||}{} & \multicolumn{1}{c|}{A-Judge}  & \multicolumn{1}{c|}{B-Judge} & \multicolumn{1}{c|}{A-Judge}  & \multicolumn{1}{c||}{B-Judge} & \multicolumn{1}{c|}{A-Judge}  & \multicolumn{1}{c|}{B-Judge} & \multicolumn{1}{c|}{A-Judge}  & \multicolumn{1}{c|}{B-Judge}  \\\hline 
Using the shortest terms  & 0.4744  & 0.4897  & 0.4488$^{*}$  & 0.4487$^{*}$  & 0.3900  & 0.5082  & 0.3850  & 0.4468\\
Using all term patterns$^{\#}$   & 0.4445  & 0.4665  & 0.4172  & 0.4180  & 0.3965  & 0.4981  & 0.3960  & 0.4444\\
Using the lattice structure    & 0.4711  & 0.4884  & 0.4436  & 0.4448  & 0.4009  & 0.5069  & 0.3884  & 0.4469 \\
Using down-weight ($k_{down} = 0.01$) & 0.4760  & 0.4896  & 0.4492  & 0.4494  & 0.3940  & 0.5082  & 0.3850  & 0.4470\\
Using down-weight ($k_{down} = 0.1$)  & 0.4816  & 0.4986  & 0.4545  & 0.4568  & 0.4003  & 0.5076  & 0.3860  & 0.4498\\\hline
\end{tabular}

\begin{minipage}[h]{14cm}
The method tagged with ``$\#$'' is a base method for comparison. 
A result tagged with ``$*$'' is superior 
to the base method's at the significance level of 5\%, 
and a result tagged with ``$**$'' is superior 
at the significance level of 1\%. 
\end{minipage}

\end{center}
\end{table*}

\section{Experiments}
In this section we describe 
several experiments done 
to test the effectiveness of 
the several methods used in our system. 
In the experimental results of this section, 
we also show 
Average Precision (the average of the precision 
when each relevant document is retrieved) 
in addition to R-Precision. 
For the comparison experiments, 
t-test is used. 
A method tagged with ``$\#$'' 
in Tables \ref{tab:result_keyward} to \ref{tab:result_detail}) 
is the base for comparison. 
A method tagged with ``$*$'' is superior 
to the base method at the significance level of 5\%, 
and a method tagged with ``$**$'' is superior 
at the significance level of 1\%. 
T-test is used only in formal-run experiments. 
(The preliminary-run data contained six queries, and 
the formal-run data contained thirteen queries.)

\subsection{Comparison of term extraction methods}

We showed the following 
four term extraction methods in Section \ref{sec:extract_keyword}. 

\begin{enumerate}
\item 
{Method using the shortest terms}
\item 
{Method using all the term patterns}
\item 
{Method using a lattice}
\item 
{Method using down-weighting}
\end{enumerate}

All the comparison results are shown in Table \ref{tab:result_keyward}. 
In Table \ref{tab:result_keyward}(a) 
all extended terms were used. 
In Table \ref{tab:result_keyward}(b) 
no extended terms were used. 
In the down-weighting method 
we tested the two cases of 
$k_{down} = 0.1$ and $k_{down}= 0.01$. 

The precision of the all-term patterns method 
was lowest in the formal run. 
It needed to be normalized using the adhoc equation. 
Since it had the lowest precision, 
it was thought to be inferior to the other methods. 
Also, it was shown by t-test 
to be significantly inferior to the shortest terms method. 

Although the down-weighting method obtained 
the highest precision 
when no extended terms were used, 
it was not as effective 
when all the extended terms were used. 
Since it was significantly different from any of the other methods, 
cannot say that it is very reliable. 
But, in the case where 
a small amount of retrieval information was used 
(i.e. no extended terms) it was very effective. 

Since only the shortest terms method is 
significantly different from 
the all-term patterns method, 
we think it is a sound method 
which can provide reliable results. 
Since the lattice and 
the down-weighting methods are not 
significantly different from 
the all-term patterns method, 
we think that they must have some problems. 
One problem that occurred when using the 
lattice method was that 
the terms used in retrieval easily changed depending on the context, 
while the down-weighting method's problem was 
that it uses the extra terms even if it down-weights them. 
However, it is thought by us that 
using longer terms in addition, is better than 
using only the shortest terms. 
We have to continue the investigation of 
term extractions.

\begin{table*}[t]
\small
  \caption{Comparison of extended numerical terms}
  \label{tab:result_hokyoukou}
  \begin{center}
\begin{tabular}[c]{|c@{ }c@{ }c||l|l|l|l||l|l|l|l|}
\multicolumn{11}{c}{(a) Comparison when using the lattice method}\\\hline
&& & \multicolumn{4}{c||}{Formal run} & \multicolumn{4}{c|}{Preliminary run} \\\cline{4-11}
\multicolumn{3}{|c||}{Numerical terms} & \multicolumn{2}{c|}{R-Precision} & \multicolumn{2}{c||}{Average precision} & \multicolumn{2}{c|}{R-Precision} & \multicolumn{2}{c|}{Average precision} \\\hline
$K_{\tiny location}$&$K_{\tiny category}$&$K_{\tiny detail}$& \multicolumn{1}{c|}{A-Judge}  & \multicolumn{1}{c|}{B-Judge} & \multicolumn{1}{c|}{A-Judge}  & \multicolumn{1}{c||}{B-Judge} & \multicolumn{1}{c|}{A-Judge}  & \multicolumn{1}{c|}{B-Judge} & \multicolumn{1}{c|}{A-Judge}  & \multicolumn{1}{c|}{B-Judge}  \\\hline 
yes&yes&yes    & 0.5031 & 0.5161 & 0.4888$^{*}$ & 0.4745 & 0.4495 & 0.5471 & 0.4625 & 0.5202\\
yes&yes&no    & 0.4764 & 0.4935 & 0.4619 & 0.4375 & 0.4092 & 0.5086 & 0.4207 & 0.4624\\
yes&no&yes    & 0.4926 & 0.5119 & 0.4808$^{*}$ & 0.4698 & 0.4599 & 0.5499 & 0.4638 & 0.5170\\
no&yes&yes    & 0.4998$^{*}$ & 0.5301$^{**}$ & 0.4731$^{*}$ & 0.4856$^{**}$ & 0.4421 & 0.5618 & 0.4383 & 0.5171\\
yes&no&no    & 0.4932 & 0.4984 & 0.4735$^{*}$ & 0.4519 & 0.4208 & 0.5083 & 0.4326 & 0.4638\\
no&yes&no    & 0.4931 & 0.5084$^{*}$ & 0.4654$^{*}$ & 0.4634$^{*}$ & 0.4085 & 0.5134 & 0.3945 & 0.4554\\
no&no&yes    & 0.4979$^{*}$ & 0.5277$^{**}$ & 0.4673$^{*}$ & 0.4829$^{**}$ & 0.4407 & 0.5603 & 0.4391 & 0.5127\\
no&no&no$^{\#}$  & 0.4711 & 0.4884 & 0.4436 & 0.4448 & 0.4009 & 0.5069 & 0.3884 & 0.4469\\\hline
\multicolumn{11}{c}{}\\
\multicolumn{11}{c}{(b) Comparison when using the shortest terms method}\\\hline
&& & \multicolumn{4}{c||}{Formal run} & \multicolumn{4}{c|}{Preliminary run} \\\cline{4-11}
\multicolumn{3}{|c||}{Numerical terms} & \multicolumn{2}{c|}{R-Precision} & \multicolumn{2}{c||}{Average precision} & \multicolumn{2}{c|}{R-Precision} & \multicolumn{2}{c|}{Average precision} \\\hline
$K_{\tiny location}$&$K_{\tiny category}$&$K_{\tiny detail}$& \multicolumn{1}{c|}{A-Judge}  & \multicolumn{1}{c|}{B-Judge} & \multicolumn{1}{c|}{A-Judge}  & \multicolumn{1}{c||}{B-Judge} & \multicolumn{1}{c|}{A-Judge}  & \multicolumn{1}{c|}{B-Judge} & \multicolumn{1}{c|}{A-Judge}  & \multicolumn{1}{c|}{B-Judge}  \\\hline 
yes&yes&yes  & 0.5012 & 0.5205$^{*}$ & 0.4935$^{**}$ & 0.4764  &0.4412  &0.5442  &0.4546  &0.5151\\
yes&yes&no  & 0.4867 & 0.4976 & 0.4704$^{*}$ & 0.4464  &0.4126  &0.5136  &0.4220  &0.4649\\
yes&no&yes  & 0.5017 & 0.5094 & 0.4850$^{*}$ & 0.4740  &0.4410  &0.5517  &0.4556  &0.5094\\
no&yes&yes  & 0.4991 & 0.5264$^{**}$ & 0.4759$^{*}$ & 0.4841$^{**}$  &0.4213  &0.5616  &0.4340  &0.5095\\
yes&no&no  & 0.4883 & 0.4952 & 0.4647$^{*}$ & 0.4444  &0.4247  &0.5076  &0.4200  &0.4614\\
no&yes&no  & 0.4824$^{*}$ & 0.4990$^{*}$ & 0.4537 & 0.4509  &0.3927  &0.5119  &0.3901  &0.4517\\
no&no&yes  & 0.4970 & 0.5242$^{**}$ & 0.4693$^{*}$ & 0.4804$^{*}$  &0.4198  &0.5595  &0.4332  &0.5070\\
no&no&no$^{\#}$  & 0.4744 & 0.4897 & 0.4488 & 0.4487  &0.3900  &0.5082  &0.3850  &0.4468\\\hline
\end{tabular}
\end{center}
\end{table*}

\subsection{Effectiveness of extended terms}

Extended terms used in this paper are classified 
into the following three categories: 

\begin{enumerate}
\item $K_{\tiny location}$ (location information)

\item $K_{\tiny category}$ (category information)

\item $K_{\tiny detail}$ (detail information)

(Here, $K_{\tiny detail}$ contains 
$K_{\tiny length}=\frac{length}{length + \Delta}$ 
which is the numerical term for a document length 
in Eq (\ref{eqn:score}).)

\end{enumerate}

\begin{table*}[t]
  \caption{Comparison of detailed numerical terms}
  \label{tab:result_detail}
  \begin{center}
\begin{tabular}[c]{|l||l|l|l|l||l|l|l|l|}\hline
& \multicolumn{4}{c||}{Formal run} & \multicolumn{4}{c|}{Preliminary run} \\\cline{2-9}
& \multicolumn{2}{c|}{R-Precision} & \multicolumn{2}{c||}{Average precision} & \multicolumn{2}{c|}{R-Precision} & \multicolumn{2}{c|}{Average precision} \\\cline{2-9}
\multicolumn{1}{|c||}{Detail terms} & \multicolumn{1}{c|}{A-Judge}  & \multicolumn{1}{c|}{B-Judge} & \multicolumn{1}{c|}{A-Judge}  & \multicolumn{1}{c||}{B-Judge} & \multicolumn{1}{c|}{A-Judge}  & \multicolumn{1}{c|}{B-Judge} & \multicolumn{1}{c|}{A-Judge}  & \multicolumn{1}{c|}{B-Judge}  \\\hline 
Neither$^{\#}$   & 0.4744 & 0.4897 & 0.4488 & 0.4487  & 0.3900  & 0.5082  & 0.3850  & 0.4468\\
$K_{\tiny descr}$ only  & 0.4878  & 0.5125$^{**}$  & 0.4614$^{*}$  & 0.4674$^{**}$  & 0.4136  & 0.5336  & 0.3930  & 0.4635\\
$K_{\tiny proper}$ only  & 0.4746 & 0.4940 & 0.4481 & 0.4523  & 0.4031  & 0.5330  & 0.4172  & 0.4765\\
$K_{\tiny nado}$ only  & 0.4630 & 0.4765 & 0.4384 & 0.4303  & 0.3973  & 0.5097  & 0.3859  & 0.4487\\
$K_{\tiny num}$ only  & 0.4744 & 0.4897 & 0.4488 & 0.4487  & 0.3900  & 0.5082  & 0.3847  & 0.4465\\
$K_{\tiny hira}$ only  & 0.4744 & 0.4897 & 0.4488 & 0.4487  & 0.3942  & 0.5074  & 0.3854  & 0.4470\\
$K_{\tiny neg}$ only   & 0.4874 & 0.5037$^{*}$ & 0.4603 & 0.4628$^{*}$  & 0.4019  & 0.5134  & 0.3967  & 0.4554\\
$K_{\tiny stopword}$ only   & 0.4713 & 0.4941 & 0.4507 & 0.4548$^{**}$  & 0.3968  & 0.5295  & 0.3985  & 0.4629\\
$K_{\tiny length}$ only  & 0.4775 & 0.4880 & 0.4472 & 0.4492  & 0.3945  & 0.5038  & 0.3809  & 0.4448\\\hline
\end{tabular}
\end{center}
\end{table*}

In order to verify the effectiveness of the above 
three extended terms, 
we carried out eight experiments in which 
these three terms were alternately used or not used. 
These experiments were performed using 
``the lattice method'' and 
``the shortest terms method''. 
The results are shown in Table \ref{tab:result_hokyoukou}. 

The last line of the table is the case 
where no extended terms were used and 
the first line of the table is the case 
where they were all used. 
When we compared the two lines, 
we found there was 
an improvement of 0.027 to 0.045 
when our extended terms were used. 
(For example, 
the average precision of A-Judgement of 
the shortest terms method 
improved from 0.4488 to 0.4935, i.e., 0.0447.) 
This indicates that 
the extended terms used in our experiment were 
totally effective. 
Retrieval precision can be improved by using 
location and category information 
in addition to Robertson's probabilistic retrieval method. 

A method that uses one of the extended terms 
is more precise than one using no extended terms. 
Thus, each extended term become effective. 
The results of the t-test show that 
each extended term has 
a significant difference in at least one evaluation criterion. 
This indicates that 
location and category information 
are independently effective. 

The main point of our paper is to 
prove that location information and category information can 
improve the precision of 
Robertson's probabilistic information retrieval method. 
This was confirmed by our experimental results. 

Use of location information is apt to 
decrease the precision of B-Judgement. 
This is because 
B-Judgement judges that 
``a document whose topic 
is partly relevant to a query'' 
is a relevant document. 
Location information weights 
a term which is in the title or at the beginning 
of the body of a document, 
i.e., a term which indicates the topic of a document. 
Therefore, for a document where 
the content of a query is written someplace 
than the topic part is not likely to be retrieved. 
The T-test also showed that 
location information is not significantly  different 
in B-Judgement. 

\subsection{Effectiveness of detail terms}

This section examines the effectiveness of 
the terms of $K_{\tiny detail}$ and 
$K_{\tiny length}$. 
In our experiments, 
the shortest terms method is used for 
term extraction. 
The values of the constants of the detail terms are 
set as in System B of Section \ref{sec:contest}. 
A comparison of the experimental results is shown 
in Table \ref{tab:result_detail}. 
The four terms 
$K_{\tiny nado}$, $K_{\tiny num}$, 
$K_{\tiny hira}$, and $K_{\tiny stopword}$ 
did not improve precision, while 
$K_{\tiny descr}$ and $K_{\tiny neg}$ 
improved precision greatly. 
This indicates that the following were confirmed by experiments: 
\begin{itemize}
\item 
  A term which is obtained from a title of a query 
  (DESCRIPTION) is important.
\item 
  A term which is obtained from a expression 
  tagged with ``NEG'' should be removed. 
\end{itemize}

\section{Conclusion}
Our information retrieval method uses 
Robertson's 2-poisson model \cite{2poisson}, which is one kind of 
probabilistic approach. But, this method 
does not use location or category information, 
which should be used to facilitate information retrieval. 
Against this background, 
we constructed a framework by using 
location, category and detailed information in a 2-poisson model. 
For the 1999 IREX contest, 
we submitted our two systems where 
their precision in the A-judgement measure was 
0.4926 and 0.4827, 
the highest values among the 15 teams and 22 systems 
in the IREX contest. 
These results indicate that 
our method is comparatively good. 

We carried out comparison experiments 
in order to confirm the effectiveness of 
each method used in our systems. 
We found that 
location and category information 
are effective 
while 
even the shortest terms method can obtain high precision. 
Also, we found several detailed facts such as 
an expression tagged with ``NEG'', should be removed. 

After this work, by using the technique of IR, 
we are conducting the research on question answering system \cite{qa_memo}. 

\section*{Acknowledgments}
We would like to thank Dr. Naoto Takahashi 
of ETL in Japan for his comment on use of category information. 
In this work, we use a lot of data of IREX.   
We would like to thank the staff and participants 
at the IREX context \cite{irex1_Sekine_eng,irex1_IR_eng,murata_irex_ir_nlp_eng}.


\bibliographystyle{plain}

\end{document}